# Research on vehicle detection based on improved YOLOv8 network

Haocheng Guo[1,4], Yaqiong Zhang[2,5], Lieyang Chen[3,6], Arfat Ahmad Khan[1,7]

[1]Khon Kaen University, Khon Kaen, Thailand
[2]University of Michigan, Ann Arbor, USA
[3]Columbia University in the City of New York, New York, USA

[4]haocheng.g@kkumail.com
[5]yaqiongz@umich.edu
[6]lc3548@columbia.edu
[7]arfatkhan@kku.ac.th

**Abstract.** The key to ensuring the safe obstacle avoidance function of autonomous driving systems lies in the use of extremely accurate vehicle recognition techniques. However, the variability of the actual road environment and the diverse characteristics of vehicles and pedestrians together constitute a huge obstacle to improving detection accuracy, posing a serious challenge to the realization of this goal. To address the above issues, this paper proposes an improved YOLOv8 vehicle detection method. Specifically, taking the YOLOv8n-seg model as the base model, firstly, the FasterNet network is used to replace the backbone network to achieve the purpose of reducing the computational complexity and memory while improving the detection accuracy and speed; secondly, the feature enhancement is achieved by adding the attention mechanism CBAM to the Neck; and lastly, the loss function CIoU is modified to WIoU, which optimizes the detection box localization while improving the segmentation accuracy. The results show that the improved model achieves 98.3%, 89.1% and 88.4% detection accuracy for car, Person and Motorcycle. Compared with the pre-improvement and YOLOv9 models in six metrics such as Precision.

**Keywords:** YOLOv8; FasterNet; CBAM mechanism; WIoU

## 1. Introduction

Vehicle detection has become a research focus in the exploration of autonomous driving technology, and it has shown a pivotal role in many practical application scenarios. The core composition of this field includes the three pillar modules of environment perception, behavioral decision-making and motion control, in which vehicle detection, as a core component of environment perception, and the equally critical behavioral decision-making module, together lay the foundation for the realization of precise motion control of vehicles. Therefore, the research on efficient and accurate vehicle target detection algorithms and their application to early warning systems are of great value in promoting the progress of this field.

 Deep learning techniques have become popular in the field of target detection and have demonstrated excellent results in the vehicle recognition task of intelligent transportation systems, especially in the classification of vehicle detection algorithms relying on traffic monitoring, deep learning-based strategies are highly favored, which can be broadly categorized into the two camps of the two-stage method relying on the candidate region and the single-stage method based on regression. Aiming at the challenge of inefficient detection of tiny vehicles in UAV images, Song et al. effectively improved the detection accuracy by integrating FPN to the Faster R-CNN framework and combining a multi-feature fusion strategy and a focal loss function[1]. Songjiang Li et al. introduced an optimized version of Cascade RCNN with an improved feature pyramid structure, which allows shallow information to gradually penetrate into the deeper layers and enhances the capture of tiny and occluded vehicle features[2].

On the other hand, Song et al. innovatively designed the MEB-YOLO model, which facilitates complex feature fusion through BiFPN architecture while controlling the computational overhead and utilizes the decoupled head to accomplish accurate prediction[3]. Focusing on improving tracking robustness, Zhang et al. introduced a high-resolution Siamese network design to address the limitations of traditional Siamese networks in bounding box estimation, and optimized the learning of target-aware masks by combining multiple attention mechanisms, which significantly improved the accuracy of vehicle recognition and ensured real-time processing capability[4][5].

Existing algorithms are unable to achieve an overall improvement between the accuracy of vehicle detection, model size and detection speed. For this reason, an improved YOLOv8 model is proposed in this paper. Specifically, using the YOLOv8n-seg model as the base model, firstly, the FasterNet network is used to replace the backbone network, which achieves the purpose of reducing the computational complexity and memory while improving the detection accuracy and speed; secondly, the feature enhancement is achieved by adding the attention mechanism CBAM to the Neck; and lastly, the loss function CIoU is modified to WIoU, which optimizes the detection frame localization while improving the segmentation accuracy.

## 2. YOLOv8 structure

The architectural essence of YOLOv8 lies in its subtle combination of advanced neural network designs, such as the adoption of CSPNet as the backbone network, which effectively reuses feature map resources and reduces the computational burden[6][7]. In addition, it embeds SPP (Spatial Pyramid Pooling) and PAN (Path Aggregation Network) modules, the inclusion of which facilitates the deep fusion of multi-scale features and greatly enhances the model's ability to capture targets of various sizes, especially the recognition accuracy of tiny targets[8].

Unlike the Anchor-Based design of previous YOLO versions, YOLOv8 adopts an Anchor-Free strategy to directly predict the geometric centers and dimensions of the bounding box, a change that simplifies the model complexity and improves the localization accuracy[9][10]. At the same time, the model utilizes the IoU loss function to fine-tune the bounding box prediction to ensure that the prediction results fit the actual target.

At the optimization and training level, YOLOv8 employs modern optimizers such as AdamW with cosine annealing learning rate strategy and Mish activation function, which together help the model achieve fast convergence and excellent generalization performance. The training process is carefully designed, starting with pre-training by freezing the backbone network, followed by full network fine-tuning to achieve optimal performance.

In the inference stage, YOLOv8 demonstrates a highly optimized process, from image preprocessing to model forward propagation to the execution of post-processes such as non-maximal suppression (NMS), each step is carefully tuned to ensure high-speed and accurate result output.

## 3. Improvements to the YOLOv8 model

*3.1. Backbone network improvements*
FasterNet, aims to increase the speed and efficiency of neural networks through an innovative partial convolution (PConv) technique that reduces redundant computations while optimizing memory access for more efficient extraction of spatial features. This design directly contributes to YOLOv8's balance between processing speed and computational resource consumption, making it an exemplary lightweight high-performance target detection algorithm.

The core features of FasterNet are the following two: Partial Convolution (PConv): Reduces unnecessary computational operations through a well-designed convolution kernel, which not only reduces the model's FLOPS (Floating Point Operations per Second), but also doesn't sacrifice the model's performance in the target detection task, realizing a win-win situation for both computational efficiency and detection accuracy. Efficient feature extraction: FasterNet is able to utilize shallow and deep features more efficiently through improved structural design, which enhances the recognition of objects at different scales, which is crucial for target detection tasks, especially when facing small targets and complex backgrounds.

The formula for calculating the size of the convolution output is given below:

$$\text{out}_{size} = \frac{\text{in}_{size} - k + 2 \times p}{s} + 1 \tag{1}$$

Where insize is the input size, k is the convolution kernel size, p is the padding size and s is the step size. Equation (1) acts with designing the network layer to ensure that the feature map size is as expected. PConv's memory usage is:

$$h \times w \times 2c_p + k^2 \times c_p^2 \approx h \times w \times 2c_p \tag{2}$$

In the above expression, h and w represent the height and width of the feature map, respectively, while k indicates the dimension of the convolution kernel, and $c_p$ identifies the number of channels on which the regular convolution actually acts. By employing the partial convolution PConv technique, the need for memory accesses plummets to one-fourth of the conventional convolution, due to the fact that the remaining c-$c_p$ channels are excluded from the computational process, thus eliminating the burden of memory accesses to these channels.

3.2.Improvements to neck

CBAM, or Convolutional Block Attention Module, is a lightweight attention mechanism module integrated into a convolutional neural network.

The channel attention mechanism aims to assign a weight to each channel of the input feature map, emphasizing the important channels and suppressing the unimportant ones. Its main steps are as follows:Global Average Pooling (GAP): First, apply global average pooling to the input feature map to obtain a vector.

Channel Attention Score Calculation: Next, channel attention scores are generated by a simple network with two fully connected (FC) layers. The first FC layer is usually followed by a ReLU activation function, and the second FC layer is followed by a sigmoid function to generate weights between 0 and 1 indicating the importance of each channel. Let the outputs of these two FC layers be v1 and v2, respectively, then the channel attention weights Mc can be expressed as:

$$v_1 = ReLU(W_1 \cdot Reshape(F_{gap}) + b_1) \tag{3}$$

$$v_2 = ReLU(W_2 \cdot Reshape(F_{gap}) + b_2) \tag{4}$$

$$M_c = \sigma(W_1 \cdot v_1 + b_1 + W_2 \cdot v_2 + b_2) \tag{5}$$

where $\sigma$ denotes the sigmoid function, $W_1$ and $W_2$ are the learnable weight matrices, and $b_1$ and $b_2$ are the bias terms.Channel weighting: Finally, the channel attention weights Mc are applied to the original feature map F to obtain the channel attention-adjusted feature map Fc:

$$F_c = M_c \odot F \tag{6}$$

where $\odot$ denotes element-by-element multiplication.

The spatial attention mechanism focuses on assigning weights at each location of the feature map, emphasizing key spatial regions in the image. Its main steps are:Maximum Pooling and Average Pooling: Firstly, Global Maximum Pooling (GMP) and Global Average Pooling (GAP) are applied to the input feature map F to obtain two vectors respectively.Feature fusion and convolution: these two vectors are stitched together to generate spatial attention weights Ms, representing the importance of each location in the feature map, by means of a convolutional layer (1×1 convolutional kernel) and a sigmoid function.

$$F_{pool} = [GMP(F); GAP(F)] \quad (7)$$

$$M_s = \sigma(Conv_{1\times 1}(Reshape(F_{pool}))) \quad (8)$$

Spatial weighting: the obtained spatial attention weights Ms are applied to the original feature map F to obtain the spatial attention adjusted feature map Fs:

$$F_s = M_s \odot F \quad (9)$$

In summary, the final CBAM expression for the attention mechanism is:

$$F_{CBAM} = F_c \odot F_s \quad (10)$$

## 4. Experiments

*4.1. DataSets and Simulation environment*

The dataset selected for this study is derived from the BDD100K automated driving dataset, which is carefully selected and constructed to cover a wide range of environmental features, including different climatic conditions and weather variations, in order to enhance the generalization ability of the model. The customized dataset focuses on the three categories of "cars", "pedestrians", and "motorcycles", and contains a total of 30,000 images and their accurate annotated files, ensuring rich resources for scene understanding and object recognition training.

Hardware environment: processor in Intel ® Core TM i7-7500U CPU @ 2.70 GHz 2.90 GHz, memory 12 GB; software environment: Microsoft Windows 10 (64 bit). The simulation software is Python 3.10.

*4.2. Vehicle test results*

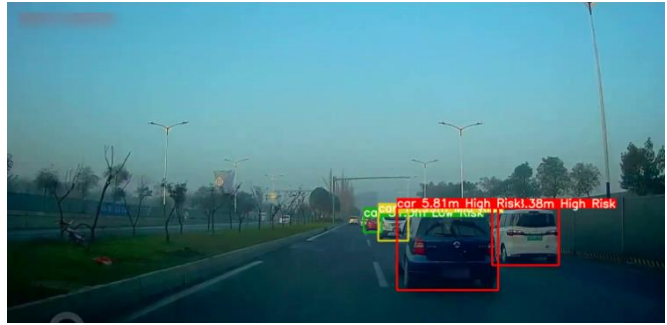

**Fig.1** Vehicle Inspection Schematic

It can be found through Fig. 1 that vehicle detection can be performed in the dataset using the improved YOLO v8 model. Next, this paper designs relevant evaluation indexes to quantitatively analyze the model.

The detection results of the improved YOLOv8 model are given next:

**Table 1 Detection of three types of targets**

| Category | Improved YOLOv8 | YOLOv8 | YOLOv9 |
|---|---|---|---|
| car | 0.983 | 0.949 | 0.962 |
| Person | 0.891 | 0.834 | 0.904 |
| Motorcycle | 0.884 | 0.832 | 0.877 |

As can be seen in Table 1, the improved network improves the detection of each type of target by up to 5.7%.

In this study, the experimental results were comprehensively evaluated by six key performance indicators, specifically Precision, Recall, Precision-Recall curve analysis, F1 score, Model Size, and Computational Cost. This multi-dimensional evaluation system is designed to examine the performance, efficiency and usefulness of the model in all aspects.

**Table 2 Comparison of 6 evaluation indicators**

| network model | Precision | Recall | PR | F1 | Size(MB) | Computation |
|---|---|---|---|---|---|---|
| Improved YOLOv8 | 0.987 | 0.99 | 0.893 | 0.85 | 8.3 | 3579152 |
| YOLOv8 | 0.954 | 0.99 | 0.835 | 0.79 | 13.5 | 4351753 |
| YOLOv9 | 0.972 | 0.99 | 0.881 | 0.81 | 11.2 | 3996254 |

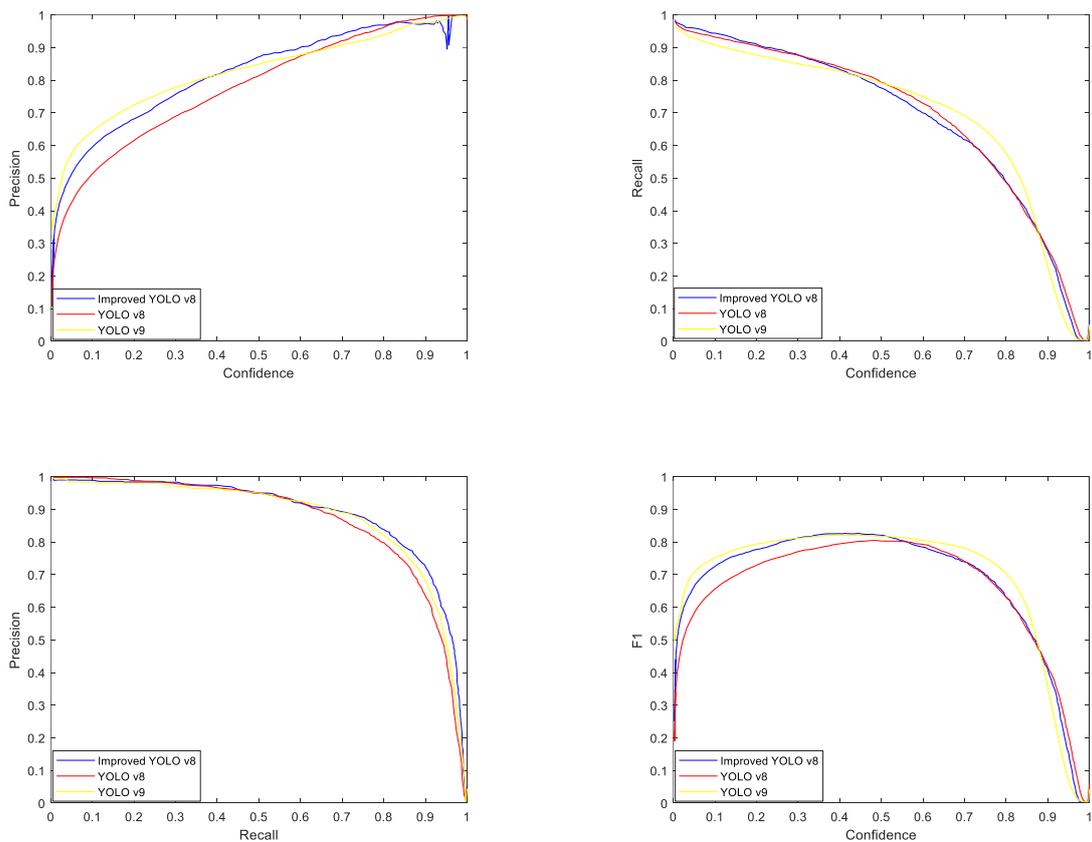

**Fig.2** Comparison charts for the 4 evaluation indicators

From Table 2 and Fig. 2, it can be concluded that the improved YOLOv8 network model has improved the check accuracy Precision by 3.46% compared to the v8 model and 1.54% compared to the v9 model. And PR curves, F1 values, model size and computation are also improved

## 5. Conclusions

This paper presents an improved YOLOv8-based vehicle detection method, which incorporates several modifications to enhance its performance. First, the FasterNet network replaces the original backbone network to improve feature extraction. Second, the attention mechanism CBAM is introduced in the Neck to refine feature maps for more accurate detection. Finally, the loss function CIoU is modified to WIoU, further improving the model's performance. Experimental results show that the improved model achieves detection accuracies of 98.3%, 89.1%, and 88.4% for cars, persons, and motorcycles, respectively. Additionally, the model demonstrates significant improvements in six performance metrics, including Precision, compared to the pre-improvement version and the YOLOv9 model.

Despite the promising results, there are several areas for future improvement. One challenge is enhancing the model's robustness in complex environments with diverse lighting conditions and occlusions. Further optimization of the backbone network and attention mechanisms could contribute to improving detection accuracy under such conditions. Additionally, reducing model inference time remains crucial for real-time vehicle detection applications. Future work will also explore the integration of the model with advanced sensor data, such as LiDAR, to further improve detection performance in various weather conditions. By addressing these challenges, the proposed method can be adapted for broader real-world applications, including smart city traffic monitoring and autonomous driving systems.